\documentclass[a4paper, 10pt, conference]{ieeeconf}      
\usepackage{FG2023}

\FGfinalcopy 

\IEEEoverridecommandlockouts                              
\usepackage{times}
\usepackage{epsfig}
\usepackage{graphicx}
\usepackage{amsmath}
\usepackage{amssymb}
\usepackage{bm}
\usepackage{fixltx2e}
\usepackage{placeins}
\usepackage{gensymb}
\usepackage{tablefootnote}
\usepackage{xcolor}
\usepackage[multiple]{footmisc}
\usepackage[font=footnotesize]{caption}
\usepackage[pagebackref=true,breaklinks=true,letterpaper=true,colorlinks,bookmarks=false]{hyperref}
\usepackage[colorlinks]{hyperref}

\def\FGPaperID{****} 

\begin{document}

\title{Analyzing the Impact of Shape \& Context on the Face Recognition Performance of Deep Networks\thanks{*Work done while at Notre Dame}}

\author{\parbox{16cm}{\centering
    {\large Sandipan Banerjee*$^1$, Walter Scheirer$^2$, Kevin Bowyer$^2$, and Patrick Flynn$^2$}\\
    {\normalsize
    $^1$ Samsung Research America, $^2$ University of Notre Dame\\
    }}
}

\maketitle

\begin{abstract}
In this article, we analyze how changing the underlying 3D shape of the base identity in face images can distort their overall appearance, especially from the perspective of deep face recognition. As done in popular training data augmentation schemes, we graphically render real and synthetic face images with randomly chosen or best-fitting 3D face models to generate novel views of the base identity. We compare deep features generated from these images to assess the perturbation these renderings introduce into the original identity. We perform this analysis at various degrees of facial yaw with the base identities varying in gender and ethnicity. Additionally, we investigate if adding some form of context and background pixels in these rendered images, when used as training data, further improves the downstream performance of a face recognition model. Our experiments demonstrate the significance of facial shape in accurate face matching and underpin the importance of contextual data for network training.
\end{abstract}

\section{Introduction}
Deep learning \cite{DLNature} models have made significant contributions to face recognition research culminating in state-of-the-art performance \cite{VGG,LightCNN,arcface} on extremely challenging datasets like MegaFace \cite{MegaFace}, IJB-B \cite{IJBB} and IJB-C \cite{ijbc}. Such high accuracy numbers are typically generated by models trained on millions of images that belong to $\sim$500K identity classes \cite{celeb500k,lwc_iccvw}. However, many of these training datasets are either private \cite{Google_FaceNet,Facebook_Deepface} or have been decommissioned \cite{exposeai} due to privacy issues \cite{MSCeleb,VGGFace2,MegaFace} and license misuse \cite{Lessons1000}. A potential solution to bypass this issue but still inject diversity to model training is to artificially augment \cite{FRAugSurvey} an existing training set. One such popular augmentation technique is to introduce graphically generated renderings of face images using 3D face models (\emph{i.e.} texture wrapping) into the downstream recognition model training \cite{HassFront,masi_ijcv,SREFI2}. Such graphical renderings, when used as supplemental training data, can present variations in facial pose and expression of existing \cite{masiFG17} or synthetic identities \cite{SREFI1,SREFI2} to a recognition model and help it learn robust representations, boosting its test performance \cite{masi_ijcv,SREFI2}.

While this augmentation approach has been shown to improve network models, it does not essentially preserve the underlying shape of the base identity as the 3D models are either randomly chosen \cite{HassFront,masiFG17} or selectively picked based on a fit-estimation function \cite{SREFI2}. The generated renderings typically look realistic but a discussion on how much of an appearance change this produces for any base identity is missing from the literature. Should the base identity and the corresponding 3D face model come from the same gender and ethnic group or should the latter be picked based on some rigid filtering in the attribute space? Using two popular techniques \cite{masiFG17,SREFI2}, we try to answer this question and understand how changing the underlying shape can perturb the appearance of both real and synthetic subjects, quantified using the authentication performance of an established network model \cite{ResNet,VGGFace2}.

Additionally, due to the missing 2D $\rightarrow$ 3D correspondences outside of the internal facial mask (\emph{i.e.} convex hull of the outer facial landmark points \cite{Dlib}), these texture wrapping methods typically do not produce realistic contextual pixels (\emph{e.g.} forehead, hair, neck, clothes) or render any at all \cite{MasiAug,HassFront,SREFI2}. In such a case, the training set is augmented with novel views of the salient facial region but masked out (\emph{i.e.} zero intensity) context and background pixels. We question the optimality of this approach and artificially add contextual pixels \cite{SREFI3} to the rendered masked face images of both real and synthetic identities. When compared to the masked face images, we find the set with synthesized context and background to generate a higher performance boost when used to supplement existing face image training data \cite{CASIA}.

While we do not present a novel framework or newly collected dataset in this paper, the originality of this work lies in our experimental analysis of face image data, and associated augmentation operations, for training CNN models. Our contributions can be summarized as follows:
\begin{enumerate}
\item  We evaluate if using randomly chosen \cite{HassFront,masi_ijcv} or well-fitted \cite{SREFI2} 3D face models, when rendering 2D face images for training data augmentation, results in different realism \cite{FID} \& face matching \cite{VGGFace2} scores (\ref{sec:2d3dfiltering}).
\item We analyze how changing the underlying shape of the base identities in these images can affect the face matching performance of deep network models during authentication experiments. We perform this analysis at various levels of facial yaw with subjects varying in gender and ethnicity (\ref{sec:shape}).
\item We investigate if adding some form of contextual pixels in these artificially rendered images, when used to train a model, further improves its verification performance during testing (\ref{sec:Exp5}).
\end{enumerate}

\begin{figure*}
\centering
   \includegraphics[width=1.0\linewidth]{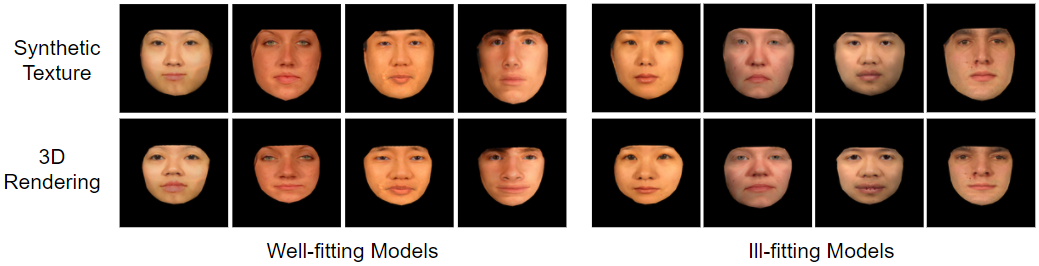}
   \caption{Sample results generated by rendering a 2D synthetic texture with well-fitting ($dist(\text{I},\text{B})$ $\leq$ $m$) and ill-fitting ($dist(\text{I},\text{B})$ $>$ $m$) 3D models for different gender and ethnicity groups. As can be seen, there is no noticeable difference in rendering quality between the two sets.}
\label{fig:2D_3D_Thres_Ex}
\vspace{-0.5cm}
\end{figure*}

\section{Related Work}
{\bf Face Recognition}: While face recognition research started with mostly handcrafted features \cite{phillips2005overview}, it has made significant progress with the advent of deep learning \cite{deepFR}. Models like FaceNet \cite{Google_FaceNet}, VGG16 \cite{VGG}, LightCNN \cite{LightCNN} and ResNet \cite{ResNet} (coupled with ArcFace loss \cite{arcface}) have achieved state-of-the-art verification performance on multiple benchmark datasets \cite{lfw,MegaFace,ijbc}. Such networks are typically trained with millions of web-scraped images, either hosted privately by commercial entities \cite{Google_FaceNet,Facebook_Deepface,masi_ijcv} or shared publicly for academic use \cite{CASIA,VGGFace2,MSCeleb}. However, many of these public datasets \cite{VGG,MSCeleb,MegaFace} have been decommissioned due to privacy related issues \cite{exposeai} and are unavailable at present \cite{Lessons1000}.

{\bf Training Data Augmentation}: To improve the diversity and coverage in training sets, researchers typically augment existing datasets before training \cite{FRAugSurvey}. Augmentation techniques include simple image transformations (\emph{e.g.} flip, rotation, scaling) to GAN based attribute translations (\emph{e.g.} modifying expression, accessories, hairstyle) \cite{starganv2,LEGAN}. While GANs can generate photorealistic face images \cite{stylegan2}, they do not always retain subject-specific traits \cite{masi_ijcv}, especially without fine-tuning during test time \cite{discofacegan}. The synthesized face images might also leak identity information from the subject set used to train the model \cite{Tinsley_2021_WACV}, which might pollute the downstream recognition task. Another approach makes use of graphics based renderings of existing faces \cite{masiFG17} or synthesized textures \cite{SREFI1} to augment the depth and width of the training set. Either generic \cite{HassFront,masiFG17} or best-fitting \cite{SREFI2} 3D face models are used to create novel views of a given facial texture by varying their facial shape, pose and expression.

{\bf Facial Shape}: Extensive research on analyzing how facial shape affects the machine, and especially human, perception of overall facial appearance has been carried out over the last two decades. \cite{InfluenceFacialFeature} analyzes how the eyes, nose, mouth and jaw features affect human judgement (\emph{i.e.} emotion, attractiveness) of face images. Relative contributions of facial shape and surface information towards the human categorization of facial expressions have been investigated in \cite{SurfaceCue}. Further studies on evaluating the race-specific face shape contributions and how they modulate neural responses have been shared in \cite{RaceShape}. As one can imagine, changes in facial shape occur frequently as people age or gain (or lose) weight. The effect of such changes in body weight and facial adiposity on the face recognition performance of different algorithms has been presented in \cite{Shape_BodyWeight,ComprehensiveStudy}. Since it affects both human and machine performance, many models have been developed that focus on learning the facial shape of individual subjects for face detection and recognition tasks \cite{PoggioShape,ASM,StatShape}. Thus, artificially augmenting the training dataset by varying the intrinsic 3D shape of a base identity can make face recognition models more robust towards such changes in unseen data \cite{masi_ijcv,SREFI2}.

{\bf Context \& Background}: Along with facial shape, the presence of contextual pixels in face images also play an important role in establishing their overall appearance. For example, external facial regions (context) are shown to be key in mapping accurate face matching by human examiners when looking at composite images \cite{Context1,Context2}. Altering these external features (\emph{e.g.} hair, ears, hooded top) in the composite images have also been shown to negatively affect face recognition performance of witnesses \cite{ExternalFeatImp,HairImp}. When internal facial features are not useful, in case of occlusion or bad lighting, human raters are shown to rely heavily on external regions for correct face matching \cite{Context3}. As demonstrated in \cite{PJP_FG}, the presence of contextual pixels also positively impacts face recognition performance of pre-trained deep networks \cite{VGG}. Preserving more contextual information when pre-processing the training and testing face images (\emph{i.e.} loose cropping) also produces better verification performance in deep learning models, especially at lower false acceptance rates \cite{DosDonts}. Leveraging the contextual information in a scene can benefit models responsible for downstream tasks like emotion recognition as well \cite{CAERNet}.

\begin{table*}
\begin{center}
\captionsetup{justification=centering}
\caption{Quantitative Analysis: Comparing the realism and face matching scores of graphical renderings of 2D face images synthesized using well-fitting 3D face models (\emph{i.e.} picked via a rigorous fit-based thresholding) and ill-fitting (\emph{i.e.} picked based on only gender and ethnicity conformity) 3D face models.}
\begin{normalsize}
\begin{tabular}{  | c| c| c| c| c|  }
\hline
{\bf Metric} & \begin{tabular}[x]{@{}c@{}}{\bf Female Asian}\end{tabular} & \begin{tabular}[x]{@{}c@{}}{\bf Female Caucasian}\end{tabular} & \begin{tabular}[x]{@{}c@{}}{\bf Male Asian}\end{tabular} & \begin{tabular}[x]{@{}c@{}}{\bf Male Caucasian}\end{tabular}\\
\hline
\hline
  FID \cite{FID} (well-fitting) & 53.19 & {\bf 42.33} & 42.94 & {\bf 46.53}\\
  \hline
  FID \cite{FID} (ill-fitting) & {\bf 52.29} & 42.48 & {\bf 42.00} & 50.31\\
  \hline
  \hline
  Similarity Score \cite{ResNet} (well-fitting) & {\bf 0.627} & {\bf 0.603} & 0.601 & 0.571\\
  \hline
  Similarity Score \cite{ResNet} (ill-fitting) & 0.592 & 0.602 & {\bf 0.618} & {\bf 0.597}\\
  \hline
\end{tabular}
\label{Tab:2D3D_Thresholding}
\end{normalsize}
\end{center}
\vspace{-0.6cm}
\end{table*}

While these works explore the importance of facial shape on the overall appearance of a human subject, they do not shed light on how much reshaping existing face images can perturb the underlying base identity. We perform an extensive set of experiments to specifically answer this question. Additionally, we estimate how the presence of contextual pixels in face images, when used for training a face recognition model, can improve its downstream verification performance.

\section{Do We Really Need Best-fitting 3D Models for Rendering Facial Texture?}\label{sec:2d3dfiltering}
To render novel views of a given face image (\emph{i.e.} 2D facial texture) the additional depth information in a 3D face model is typically utilized. While 3D data corresponding to 2D images is present in some datasets \cite{SREFIDonor,Basel}, such information is absent in popular face datasets used to train deep networks \cite{CASIA,MSCeleb}. In the absence of such depth information, researchers take advantage of proxy 3D face models available in public datasets \cite{Basel,SREFI2}. This proxy 3D model can be either generic \cite{HassFront}, randomly chosen \cite{masiFG17} or selectively picked based on some fitting function between the 3D model and the 2D image \cite{SREFI2}. In this section, we experimentally judge if a rigorous model-fitting function is really required to generate realistic renderings or can a simple gender and ethnicity based model selection approach achieve the desired result.

To approximate the compatibility of a 3D face model to a 2D face image, we use the fitting-function proposed in \cite{SREFI2}. For a given facial texture $\text{{\bf I}}_{s}$, of a synthetically generated identity in \cite{SREFI2}, a distance score $dist$ is calculated as follows:
\vspace{-0.1cm}
\begin{equation}
dist(\text{{\bf I}}_{s},\text{B}) = w_{1} * \left \| \text{{\bf v}}^{\text{{\bf I}}_{s}} - \text{{\bf v}}^{\text{B}} \right \|_1 +
w_{2} * \left \| \text{{\bf f}}^{\text{{\bf I}}_{s}} - \text{{\bf f}}^{\text{B}} \right \|_1,
\label{eq:best3D}
\end{equation}

where B is a 2D scan of the same gender and race as the base identity in $\text{{\bf I}}_{s}$ and serves as a 2D substitute for a 3D model of the same subject from the Notre Dame Synthetic Face Dataset \cite{SREFI2}. $\text{{\bf v}}^{\text{I}}$ represents the 68 landmark points of the image {\bf I}, extracted using Dlib \cite{Dlib}. $\text{{\bf f}}^{\text{I}}$ serves as the textural representation of I and is extracted from the pre-trained VGGFace network \cite{VGG}. Given this shape and textural information of each image, $dist(\text{{\bf I}}_{s},\text{B})$ captures the dissimilarity in appearance between facial texture and 3D models in these attribute spaces. In \cite{SREFI2}, the $k$ 3D models corresponding to the scans producing the lowest $dist$ values are deemed best-fitting to $\text{{\bf I}}_{s}$ and used for rendering.

For this study, we accumulate the $dist$ scores for all the 2D facial textures from the synthetic dataset released in \cite{SREFI2} and then compute the mean $m$ from this set and set it as a threshold for filtering out ill-fitting 3D models. Mathematically this can be written as:
\vspace{-0.1cm}

\begin{equation}
m = \frac{1}{kn}\sum_{i=1}^{n}\sum_{j=1}^{k}dist(\text{I}^i,\text{B}_j),
\label{eq:acc3D}
\end{equation}

where $n$ is the total number of 2D textures, $\text{I}_{i}$ is the $i$-th texture, and $\text{B}_j$ is the $j$-th best-fitting model for $\text{I}_{i}$.

The mean $m$ can be treated as an approximate threshold for 2D-3D fitness that is computed dynamically once we have all the fitness scores available. Even if $m$ is numerically different across datasets, the overall fitting behavior should not change drastically, as the associated features for each gender-ethnicity bucket and the 2D-3D fitting functions remain static. Any 3D model corresponding to a 2D scan (B) that generates a $dist$ above $m$ can be labeled as ill-fitting while still keeping the ones that generate $dist$ within $m$ as well-fitting. When enforcing a rigid goodness of fit, we can simply filter out the ill-fitting models before rendering.

\begin{figure*}
\centering
   \includegraphics[width=1.0\linewidth]{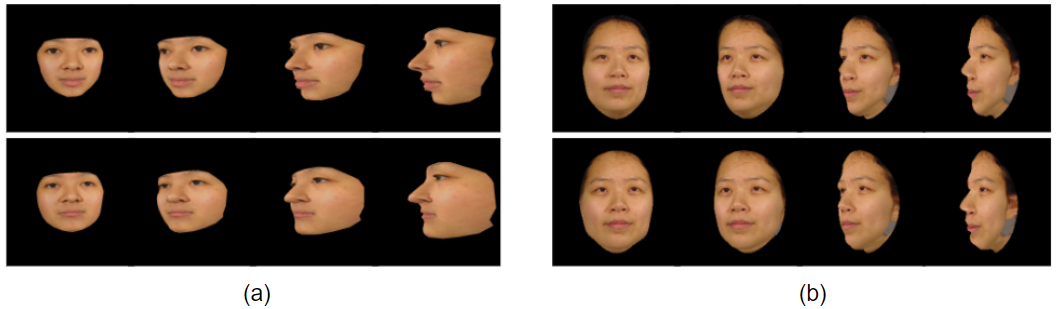}
   \caption{Sample results generated by (a) \cite{SREFI2} (synthetic identity), and (b) \cite{masiFG17} (real identity). Each row of a sub-figure shows the same facial texture (identity) rendered using a 3D face model at different facial poses. Notice how different the same facial texture looks when rendered with different 3D models (\emph{i.e.} variable shape). An analysis of how this variability in facial shape affects recognition performance can be found in \ref{sec:shape}.}
\label{fig:SREFI_Masi_Ex}
\vspace{-0.5cm}
\end{figure*}

To estimate the effect of this filtering approach, we use the following metrics: (a) FID \cite{FID} to approximate the realness of the novel views produced by 3D rendering, and (b) a similarity score between the 2D synthetic texture and the final rendered views using the pre-trained ResNet50 \cite{ResNet,VGGFace2} model to capture the structural closeness between the two. Specifically, we compute these metrics for both the well-fitting ($dist(\text{I},\text{B})$ $\leq$ $m$) and ill-fitting ($dist(\text{I},\text{B})$ $>$ $m$) 3D models for a particular 2D synthetic texture $\text{I}$. To gain a deeper perspective, we also estimate this separately for each gender-ethnicity group annotated in the synthetic dataset from \cite{SREFI2}. To focus solely on the facial region, the facial mask of the 2D synthetic texture is cropped out using the convex hull of its landmark points, as shown in Fig. \ref{fig:2D_3D_Thres_Ex}. The results are shared in Table \ref{Tab:2D3D_Thresholding}.

As can be seen, the thresholding does not dramatically improve the realism or the structural similarity of the 3D rendered views. This can be attributed to the fact that both the feature and landmark spaces used to compute $dist$ in \ref{eq:best3D} are already normalized based on gender and ethnic attributes \cite{SREFI2} and hence any best-fitting 3D model is already withing an implicit threshold for a given facial texture. Therefore, employing a simple gender-ethnicity based model selection approach, instead of an explicit fit-estimation function \cite{SREFI2}, can effectively estimate the wellness of the 2D-3D fit.

We also investigated the effect of gender and race in 2D-3D fitting by relaxing the gender-ethnicity constraint while finding compatible 3D models for a given base identity. In most cases the best fitting 3D models did belong to individuals in the same gender-ethnicity bucket as the base identity. Consequently, we added this constraint to reduce computation time.

\section{Impact of Facial Shape on Subject Identity}\label{sec:shape}
In this section, we seek to answer the following questions:
\begin{enumerate}
    \item How much of an influence does the facial shape have on a subject's identity? More specifically, if different 3D models are used to render the same facial texture, how much does the appearance change?
    \item Is the appearance change, if any, consistent at different facial poses?
    \item Is the appearance change, if any, noticeably different across varying rendering methods like \cite{masiFG17,SREFI2}?
    \item Is the appearance change, if any, similar across different genders and ethnic groups?
\end{enumerate}

As discussed before, a popular technique for augmenting training datasets is to render 2D face images with different 3D models at different facial yaw values \cite{HassFront,masiFG17,SREFI2}. It is argued that although the facial shape changes, the facial texture remains intact, and therefore the synthetic images are assigned the same identity label (an example can be seen in Fig. \ref{fig:SREFI_Masi_Ex}). These synthetic views infuse more variation in facial shape and pose and augments the depth of the training dataset. As image pairs of a test subject during authentication experiments can vary in age \cite{FGNet,MegaFace}, and consequently the facial shape \cite{Shape_BodyWeight}, these synthetic views in training can make the network more robust to such fluctuations, as shown in \cite{MasiAug,SREFI2}. However, the amount of variation in appearance this augmentation introduces is not analyzed, nor is how different the face looks overall even with the facial texture intact. To answer these questions, we randomly sample three different sets of 1,000 synthetic identities from the Notre Dame Synthetic Face Dataset \cite{SREFI2}, with each facial texture (identity) rendered with three different 3D models, at yaw values [0\degree, -30\degree, -60\degree, -90\degree]. Similarly, we sample three sets of 1,000 real identities from a larger gallery of images \cite{SREFIDonor,SREFI1}, with each facial texture (identity) rendered with three randomly chosen 3D models (of the 10 available in \cite{Basel}), at yaw values [0\degree, -22\degree, -55\degree, -75\degree]. The yaw values are picked for both sets as consistently as possible to each other.

\begin{figure}[t]
\centering
   \includegraphics[width=1.0\linewidth]{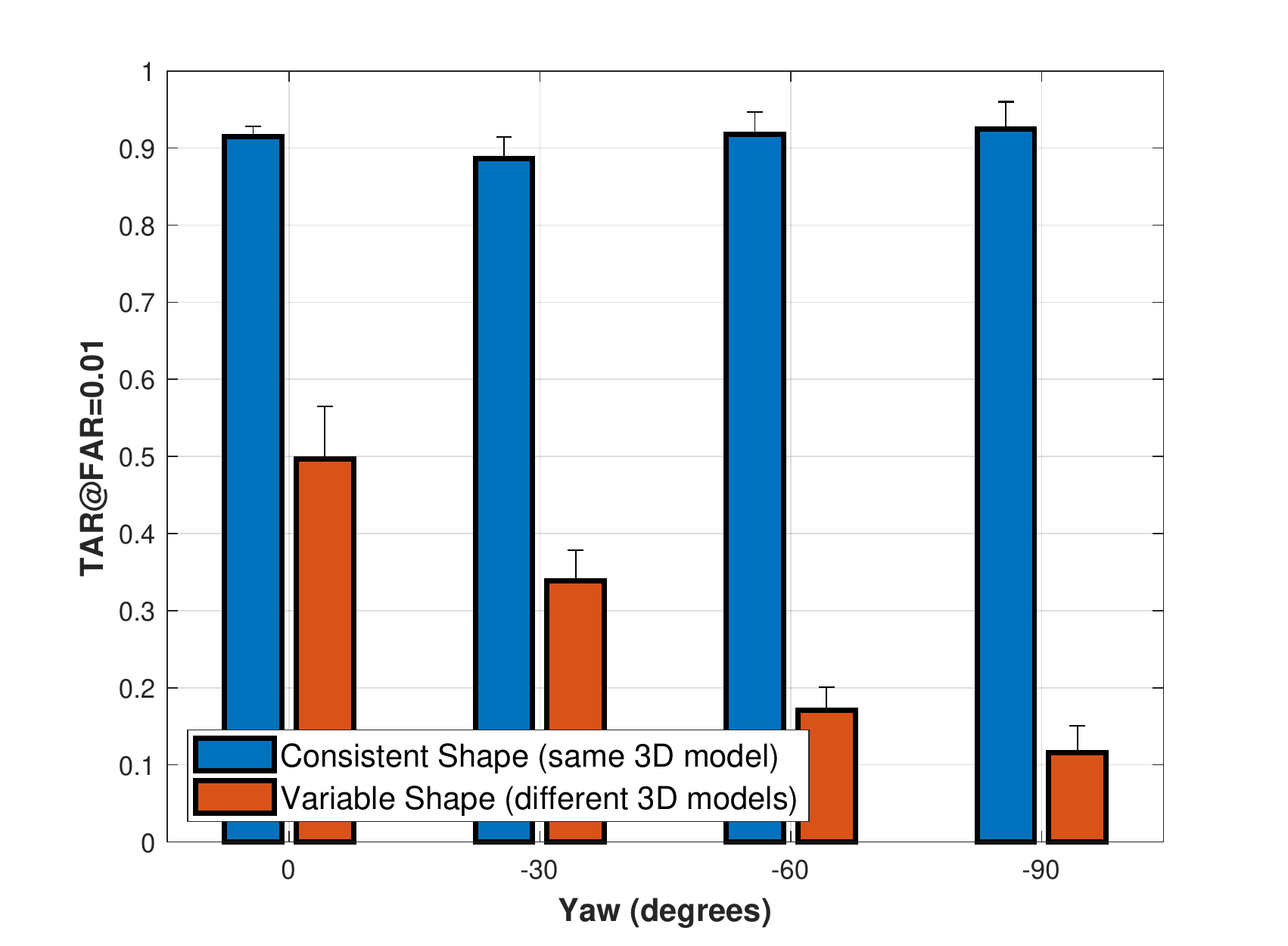}
   \caption{Verification rate with varying facial shape at different facial yaw with synthetic images generated using \cite{SREFI2}}
\label{fig:SREFI_Shape}
\vspace{-0.6cm}
\end{figure}

\begin{figure}[t]
\centering
   \includegraphics[width=1.0\linewidth]{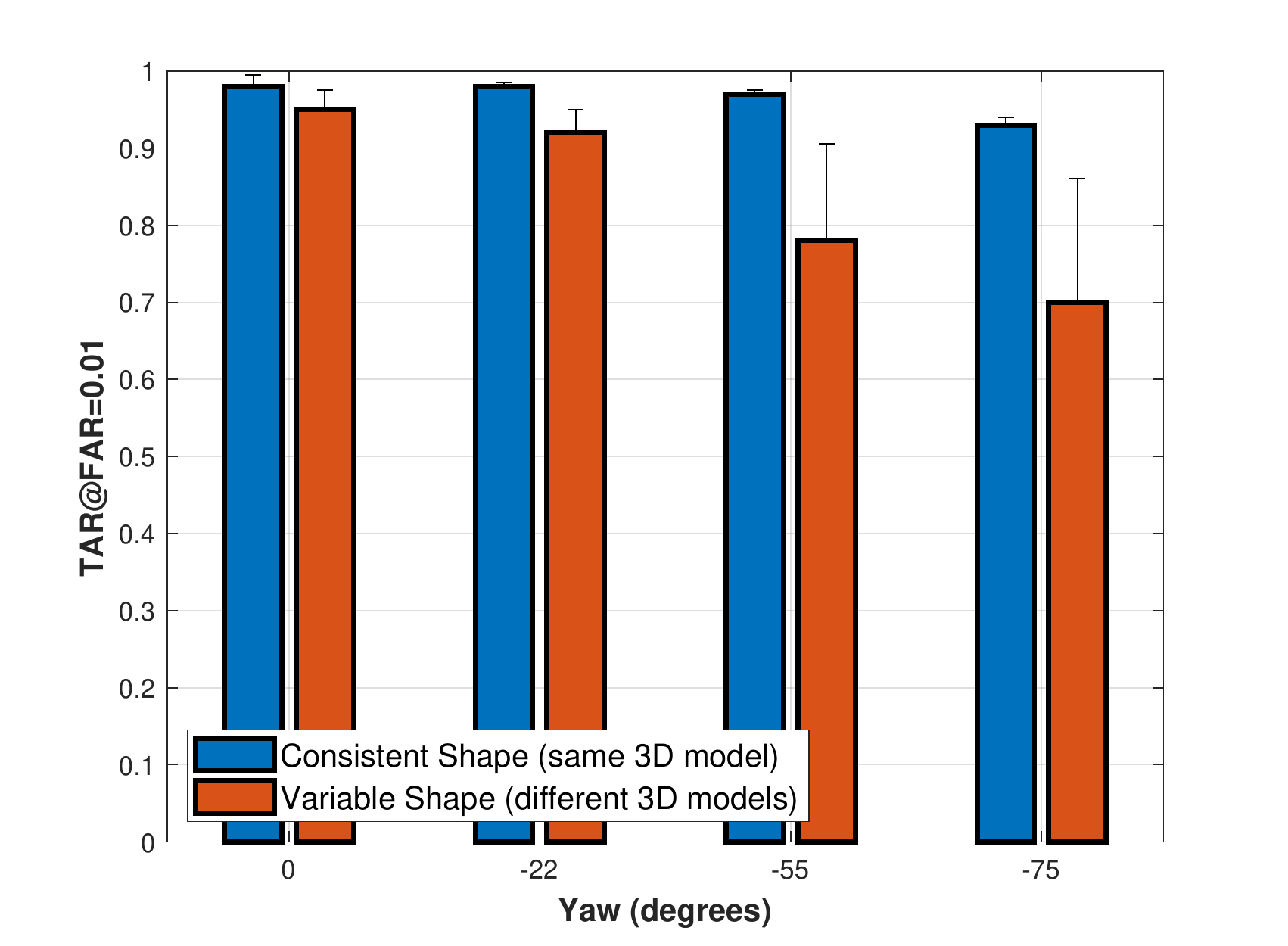}
   \caption{Verification rate with varying facial shape at different facial yaw with synthetic images generated using Masi et al's method \cite{masiFG17}}
\label{fig:Masi_Shape}
\vspace{-0.4cm}
\end{figure}

For a particular facial yaw value in each of the two datasets, we perform a verification experiment using face images (\emph{i.e.} the 2D facial textures) rendered with: (1) the same 3D model resulting in a consistent shape, and (2) different 3D models generating variable shapes. Once generated, these rendered views are passed through the ResNet50 model \cite{ResNet}, pre-trained on the VGGFace2 dataset \cite{VGGFace2}, and the 256-dimensional descriptor from the \emph{feat\_extract} layer is used as their representative features. We then use the Pearson correlation coefficient to compute match score between different feature pairs. Since each facial texture, real (Masi et al.'s method \cite{masiFG17}) or synthetic (SREFI \cite{SREFI2}), is rendered with three different 3D models, we perform the experiment separately for each 3D model. The resulting verification scores are shared in Figures \ref{fig:SREFI_Shape} and \ref{fig:Masi_Shape} respectively.

As can be seen, the verification performance drops rapidly as the facial pose moves away from 0\degree yaw. This effect is much more prominent in the SREFI generated synthetic images, compared to the ones generated by Masi et al.'s method. The reason behind this is the normalization along the jaw-line that is present in Masi et al.'s method, which makes the variation in shape less noticeable. SREFI, on the other hand, keeps the 3D model intact and therefore the variation in facial shape becomes more noticeable at higher yaws (as apparent from Fig. \ref{fig:SREFI_Masi_Ex}). It is interesting to note that for SREFI, the three 3D models chosen for rendering are the most compatible with the facial texture (using Eq.~\ref{eq:best3D}), and therefore are similar to each other to a certain degree. Even in that case, the variation in appearance between two different renderings is quite large as evident from Fig. \ref{fig:SREFI_Masi_Ex}:a and the plots in Fig. \ref{fig:SREFI_Shape}. Such a variation in facial shape, when used for training a network, is definitely beneficial to its feature learning as exhibited in \cite{MasiAug,masiFG17,SREFI1,SREFI2} and Table I in the supplementary text accompanying this manuscript.

One thing to note in regards to facial shape is that SREFI \cite{SREFI2} estimates compatible 3D face models for synthetic textures and hence the 3D renderings exhibit plausible appearance variations of the actual synthetic identity. This, in turn, boosts the recognition efficacy of a network when used for training \cite{SREFI2}. However, using generic 3D face models, as done in \cite{HassFront,MasiAug,masiFG17}, might generate implausible results (\emph{e.g.} female facial texture rendered with a male 3D face model), and make the network confuse one identity with another due to similar facial shape. Such trained networks can become more vulnerable to presentation attacks, \emph{i.e.}, distractors, with different people wearing the same facial texture mask to generate false matches \cite{Marcel_BTAS,Marcel_IWBF}.

\subsection{Gender-Ethnicity Analysis}
We further analyze the role of gender and ethnicity while estimating the impact of consistent and variable 3D facial shape on face recognition performance. For this experiment, we select synthetic identities from each of the gender-ethnic groups present in the Notre Dame Synthetic Face Dataset \cite{SREFI2} - Female Asian (FA), Female Caucasian (FC), Male Asian (MA) and Male Caucasian (MC). Since there are 593 synthetic identities altogether in the FA group, we randomly select 593 identities from the FC, MA, and MC groups, with each facial texture (identity) rendered with three different 3D models, at yaw values [0\degree, -30\degree, -60\degree, -90\degree]. It is to be noted however that each of these three 3D models are from individuals of the same gender and ethnicity (as discussed in \ref{sec:2d3dfiltering}), consistent with that of the generated synthetic texture.

\begin{figure}
\centering
   \includegraphics[width=1.0\linewidth]{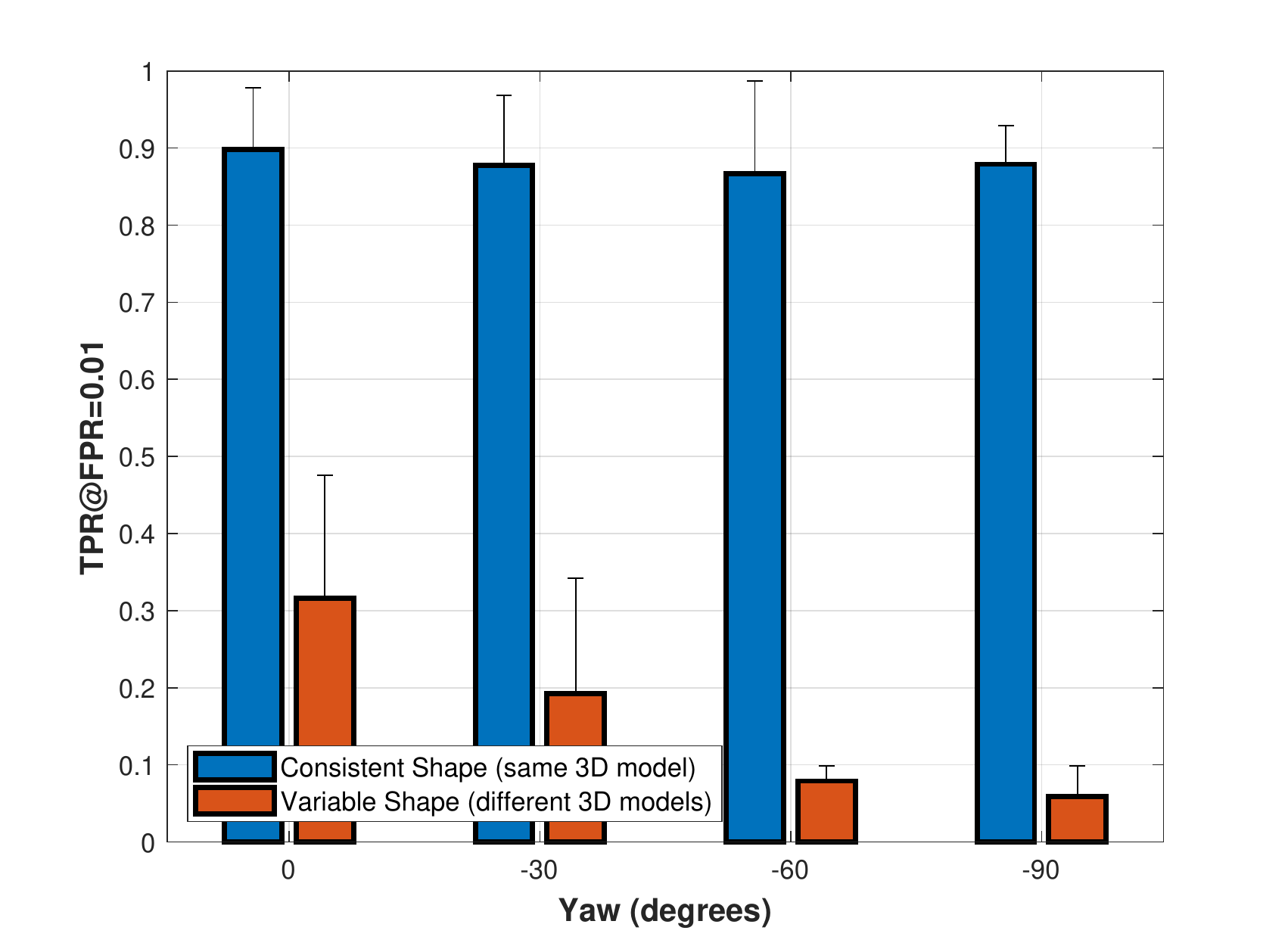}
   \caption{Verification rate with varying facial shape at different facial yaw values with Female-Asian synthetic images generated using \cite{SREFI2}.}
\label{fig:SREFI_Shape_FA}
\vspace{-0.5cm}
\end{figure}

\begin{figure}
\centering
   \includegraphics[width=1.0\linewidth]{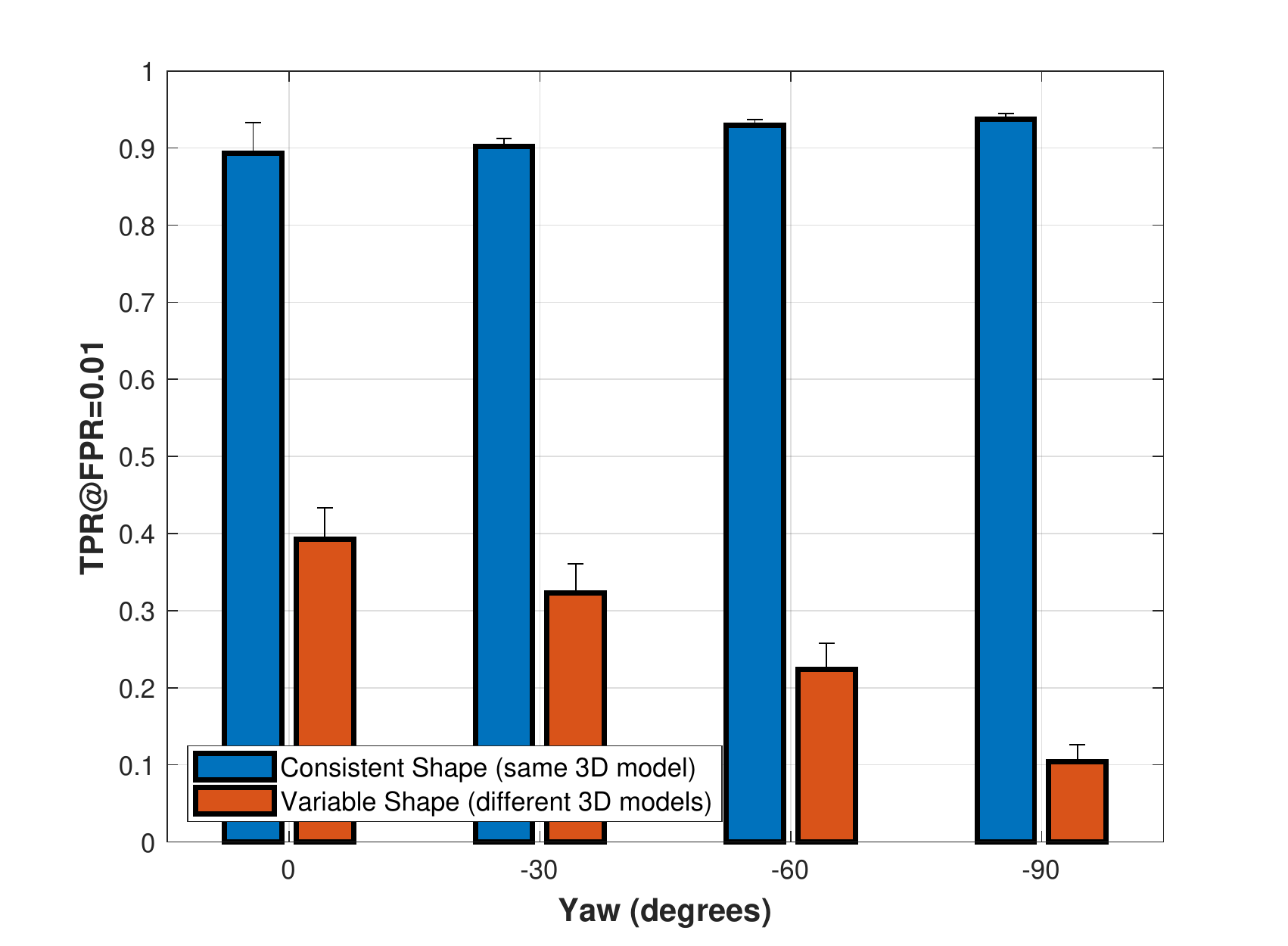}
   \caption{Verification rate with varying facial shape at different facial yaw values with Female-Caucasian synthetic images generated using \cite{SREFI2}.}
\label{fig:SREFI_Shape_FW}
\vspace{-0.5cm}
\end{figure}

\begin{figure}
\centering
   \includegraphics[width=1.0\linewidth]{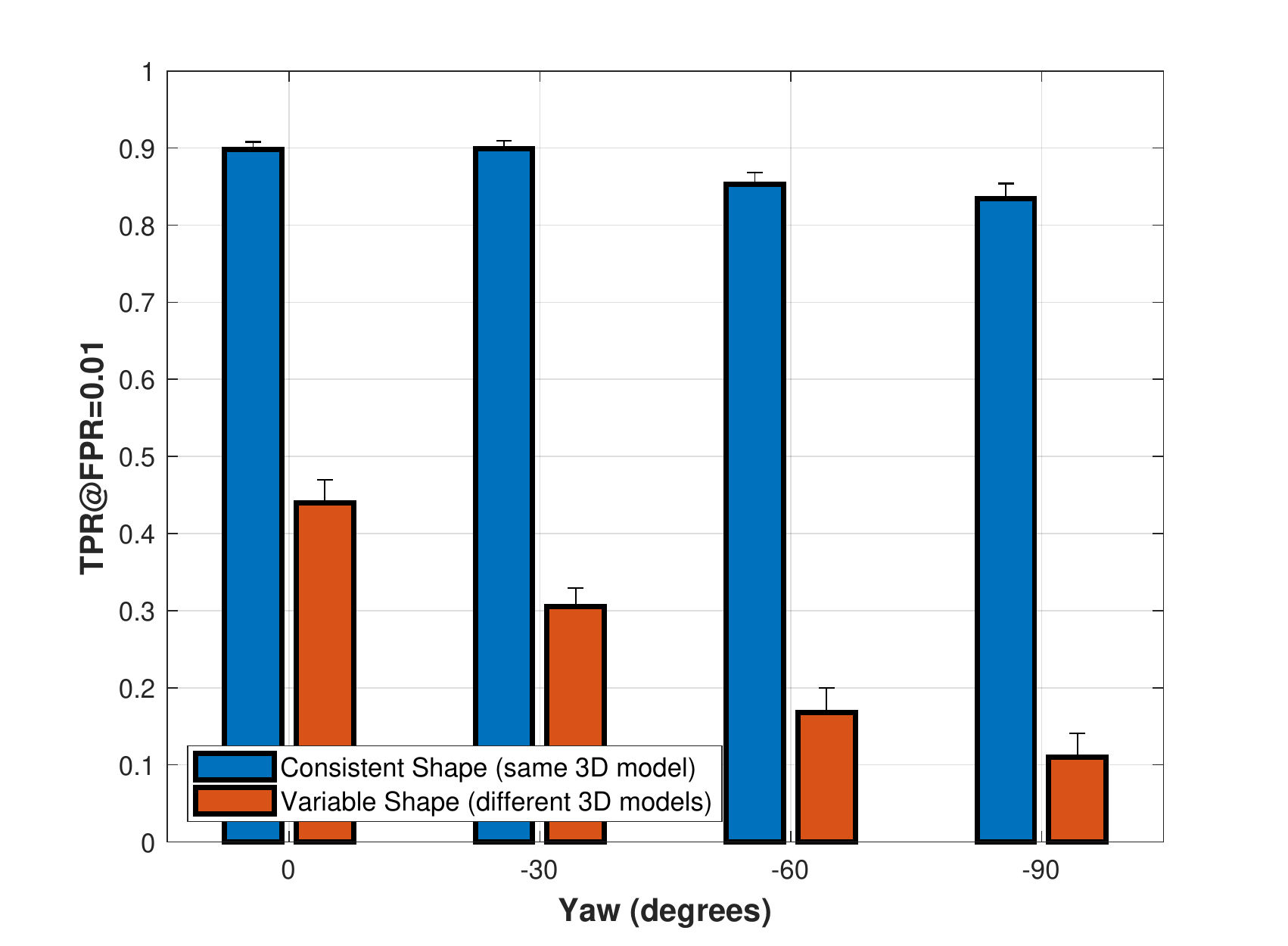}
   \caption{Verification rate with varying facial shape at different facial yaw values with Male-Asian synthetic images generated using \cite{SREFI2}.}
\label{fig:SREFI_Shape_MA}
\vspace{-0.5cm}
\end{figure}

\begin{figure}
\centering
   \includegraphics[width=1.0\linewidth]{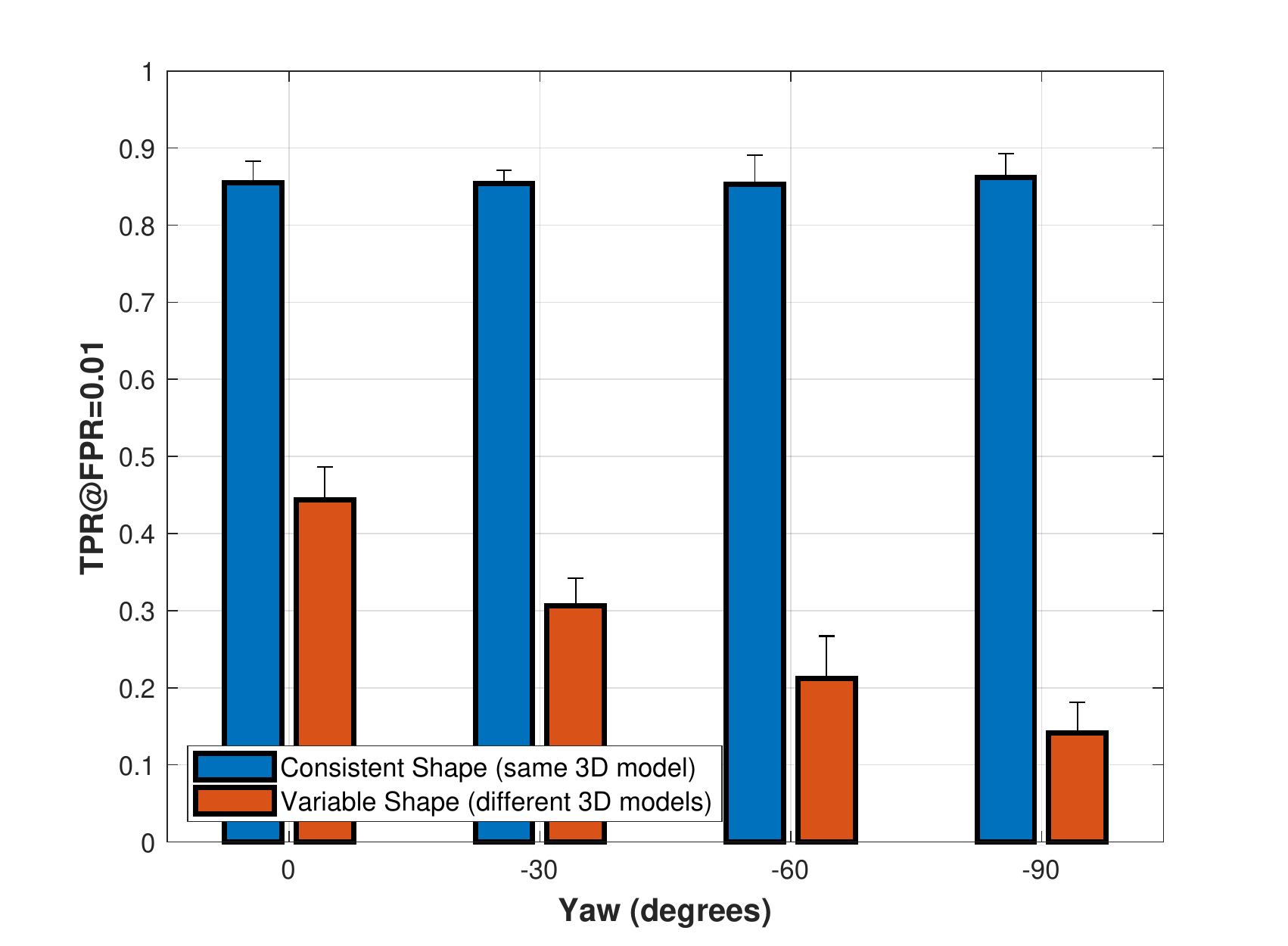}
   \caption{Verification rate with varying facial shape at different facial yaw values with Male-Caucasian synthetic images generated using \cite{SREFI2}.}
\label{fig:SREFI_Shape_MW}
\vspace{-0.5cm}
\end{figure}

For a particular facial yaw value and a certain gender-ethnic group, we perform a verification experiment with facial textures rendered with the same (consistent shape) and different 3D models (variable shape), same as before. The verification scores, resulting from this experiment, can be seen in Figures \ref{fig:SREFI_Shape_FA}, \ref{fig:SREFI_Shape_FW}, \ref{fig:SREFI_Shape_MA}, and \ref{fig:SREFI_Shape_MW}. They show a similar trend of diminishing performance when the same synthetic facial texture, rendered with different 3D models, are matched with each other. The drop in performance is even more prominent in the Asian identities (Figures \ref{fig:SREFI_Shape_FA} and \ref{fig:SREFI_Shape_MA}) than the Caucasian identities (Figures \ref{fig:SREFI_Shape_FW} and \ref{fig:SREFI_Shape_MW}), possibly due to the presence of significantly more Caucasian 3D models in the Notre Dame Synthetic Face Dataset dataset \cite{SREFI2}. With a larger set of 3D models, the chances are higher that the three 3D models, compatible to the synthetic texture, will be more closer in facial shape and generate a higher match score.


\begin{table*}
\begin{center}
\captionsetup{justification=centering}
\caption{Distribution of the training datasets created for Experiment 5. The contextual pixels for the datasets are either preserved from a frontal view \cite{masiFG17} or synthesized artificially \cite{SREFI3}.}
\begin{normalsize}
\begin{tabular}{  | c| c| c| c| c| c|  }
\hline
\begin{tabular}[x]{@{}c@{}}{\bf Training}\\{\bf Data}\end{tabular} & \begin{tabular}[x]{@{}c@{}}{\bf CASIA \cite{CASIA}}\\{\bf Images}\end{tabular} & \begin{tabular}[x]{@{}c@{}}{\bf Masi \cite{masiFG17} Images}\\(wo/ context)\end{tabular} & \begin{tabular}[x]{@{}c@{}}{\bf Masi \cite{masiFG17} Images}\\(w/ context)\end{tabular} & \begin{tabular}[x]{@{}c@{}}{\bf SREFI \cite{SREFI2} Images}\\(wo/ context)\end{tabular} & \begin{tabular}[x]{@{}c@{}}{\bf SREFI \cite{SREFI2} Images}\\(w/ context)\end{tabular} \\
\hline
\hline
  Masi\_woContext  &  \begin{tabular}[x]{@{}c@{}}111,276\\(1,452 real\\identities)\end{tabular}  & \begin{tabular}[x]{@{}c@{}}111,276\\(1,452 real\\identities)\end{tabular}  &  0  &  0 & 0\\
  \hline
      SREFI\_woContext  &  \begin{tabular}[x]{@{}c@{}}111,276\\(1,452 real\\identities)\end{tabular}  &  0  &  0  &  \begin{tabular}[x]{@{}c@{}}111,276\\(1,452 synth.\\identities)\end{tabular} & 0\\

  \hline
Masi\_Context & \begin{tabular}[x]{@{}c@{}}111,276\\(1,452 real\\identities)\end{tabular}  &  0  &  \begin{tabular}[x]{@{}c@{}}111,276\\(1,452 real\\identities)\end{tabular}  &  0 & 0\\
  \hline
      SREFI\_Context & \begin{tabular}[x]{@{}c@{}}111,276\\(1,452 real\\identities)\end{tabular}  &  0  &  0  &  0  &  \begin{tabular}[x]{@{}c@{}}111,276\\(1,452 synth.\\identities)\end{tabular}\\
     \hline
\end{tabular}
\label{Tab:Exp5}
\end{normalsize}
\end{center}
\vspace{-0.4cm}
\end{table*}

\section{Impact of Context \& Background on Face Recognition Performance}\label{sec:Exp5}
In this section, we attempt to answer these questions:

\begin{enumerate}
\item Does the presence of context (forehead, hair, neck, clothes) and background in the training data (real or synthetic) for deep networks generate a higher test performance, when compared with masked face images (\emph{i.e.} no context)?
\item In a similar vein, does the presence of context and background in the supplemental training data, generated by SREFI \cite{SREFI2} or Masi et al.'s method \cite{masiFG17}, generate a higher test score?
\end{enumerate}

For this experiment, we prepare two hybrid datasets using real face images from the CASIA-WebFace dataset \cite{CASIA} and synthetic images generated using (1) Masi et al.'s method \cite{masiFG17} and (2) SREFI \cite{SREFI2}. \cite{masiFG17} warps a given face image using a static set of ten 3D face models from the Basel dataset \cite{Basel} at pre-determined yaw values [0\degree, -22\degree, -40\degree, -55\degree, -75\degree]. Therefore, for a given face image of a real identity, it can generate 50 (10$\times$5) new views varying in facial shape and pose. However, it cannot create new identities and to generate the hybrid dataset we again use the external gallery from \cite{SREFIDonor}. It contains a total of 15,807 face images of 1,452 real subjects and using \cite{masiFG17} we render 790,350 synthetic views of the same 1,452 subjects varying facial shape and pose, and coin it as the ``Masi Dataset". These images are rendered without any context (\emph{e.g.} masked faces in Fig. \ref{fig:SREFI_Masi_Ex}.b).

Since Masi et al.'s method \cite{masiFG17} cannot generate new identities, we re-sample the real training data from CASIA-WebFace \cite{CASIA} to keep the number of subjects the same with the Masi Dataset. Specifically, we randomly select 111,276 face images of 1,452 subjects from the head, \emph{i.e.} half of the dataset containing the subjects with the most number of images of CASIA-WebFace's distribution \cite{MasiAug}. For the supplemental data generated by \cite{masiFG17}, a total of 111,276 images is selected from the 1,452 real subjects from the Masi Dataset. Similarly for SREFI \cite{SREFI2}, we sample three sets of 111,276 images of 1,452 synthetic subjects from the Notre Dame Synthetic Face Dataset \cite{SREFI2}. When combined with the CASIA-WebFace samples, these synthetic samples form a pair hybrid datasets of masked face images, as listed in the first two rows of Table \ref{Tab:Exp5}.

\begin{figure}
\centering
  \includegraphics[width=1.0\linewidth]{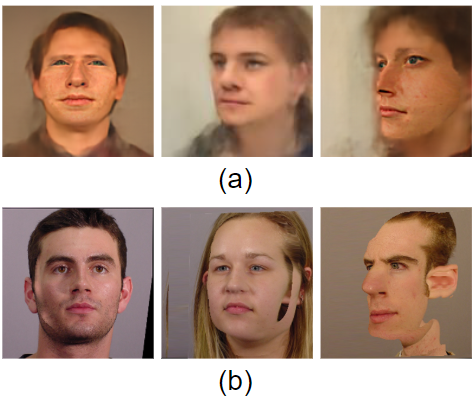}
  \caption{(a) Synthetic face images (synthetic identities) generated by SREFI \cite{SREFI2} with context and background hallucinated automatically using \cite{SREFI3}, (b) synthetic face images (real identities) generated by Masi et al.'s method \cite{masiFG17} preserving the background. All the images are 128$\times$128 in size.}
\label{fig:SREFI_Masi_wBack_Ex}
\vspace{-0.7cm}
\end{figure}

Since Masi et al.'s method \cite{masiFG17} has the functionality to preserve the context and background while rendering a frontal face image, we prepare another hybrid dataset (``Masi\_Context" in Table \ref{Tab:Exp5}) where the context of the same synthetic images are preserved. The SREFI method \cite{SREFI2} however does not render context and background pixels at non-frontal poses due to missing 2D $\rightarrow$ 3D correspondences for these external facial regions. To hallucinate the context pixels for these images, we use the trained snapshot of the multi-scale GAN model from \cite{SREFI3}. This model is composed of a cascaded network of GAN blocks, each tasked with hallucination of missing pixels at a particular resolution while guiding the synthesis process of the next GAN block. It can generate the missing pixels automatically, by taking cues from the features of the provided face mask, without requiring any human supervision. We generate context and background pixels for the 111,276 face images of the 1,452 SREFI-based synthetic identities, and prepare a hybrid dataset with sampled images from CASIA-WebFace \cite{CASIA} (``SREFI\_Context" in Table \ref{Tab:Exp5}). Sample synthetic face images with such artificially rendered context and background can be found in Fig. \ref{fig:SREFI_Masi_wBack_Ex}.

Once these datasets are prepared, we fine-tune the ResNet-50 network \cite{ResNet}, pre-trained on the VGGFace2 dataset \cite{VGGFace2}, separately in four different training sessions, using the four datasets from Table \ref{Tab:Exp5}. We use a polynomial decay policy \cite{bottou2010large} for training each network with the same batch size = 16, base learning rate = 0.001, gamma = 0.96, momentum = 0.009 and a step size of 32K training iterations. We stop training when the validation loss plateaus across an epoch.

\begin{figure}[t]
\centering
   \includegraphics[width=1.0\linewidth]{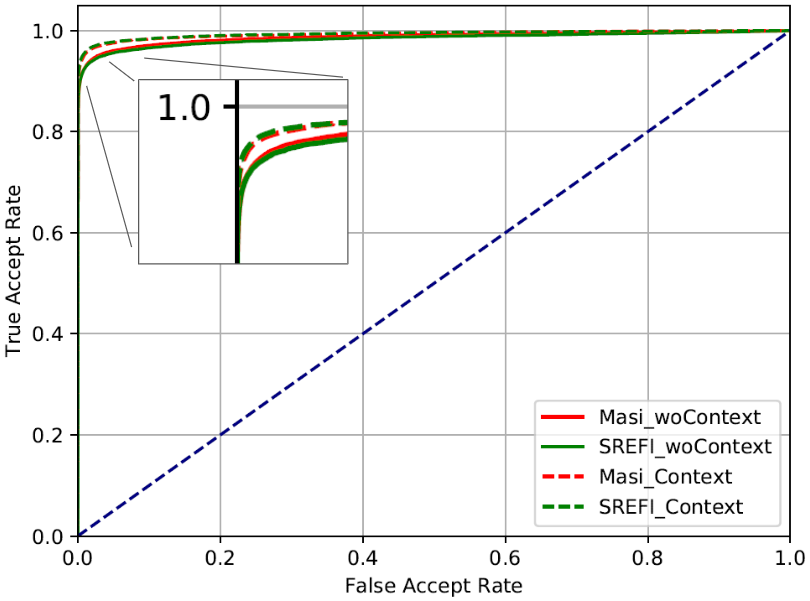}
   \caption{Verification performance of the ResNet-50 model \cite{ResNet,VGGFace2} on the IJB-B \cite{IJBB} dataset, when fine-tuned with training data with and without the presence of context and background pixels. As the curves suggest, presence of facial context benefits network training.}
\label{fig:Exp5_ROC}
\vspace{-0.7cm}
\end{figure}

For testing each trained network, we use the IJB-B verification protocol \cite{IJBB} as our performance metric. Each still image or video frame from a \emph{template} is first aligned about its eye center and then fed to each of the four trained networks for feature extraction (\emph{i.e.} 256-D descriptor from \emph{feat\_extract} layer). We generate an average feature vector for a template using \emph{video} and \emph{media pooling} operations, described in \cite{masiFG17}, and match a pair of templates using a Pearson correlation coefficient metric ($\rho$) between their feature vectors as:

\begin{equation}
\rho = \frac{Cov(F_{i},F_{j})}{\sigma_{F_{i}}\sigma_{F_{j}}},
\label{eq:Pearson}
\end{equation}
where \emph{Cov} denotes covariance, $F_{i}$ and $F_{j}$ are the pooled feature vectors of the $i$-th and $j$-th templates respectively. In an ideal situation pooled features of two templates of the same identity should match perfectly, \emph{i.e.}, $\rho$ should be close to 1. The resulting ROC curves are presented in Fig \ref{fig:Exp5_ROC}.

As the curves clearly suggest, training the network with supplemental face images containing contextual pixels makes it more robust to visual changes during testing. This can be attributed to the additional information and variation the presence of context and background provides to the network during training. Such variations in the training data actually makes the model robust to intra-class variance \cite{MasiAug} and less prone to overfitting, improving its test performance over only training with face masks. Moreover, these results are along the same lines of previous research that underpin the importance of context in face recognition by humans \cite{Context3,PJP_FG} and deep networks \cite{DosDonts}.

\section{Training Data Augmentation Results}
\vspace{-0.2cm}
Similar to the dataset preparation in Section \ref{sec:Exp5}, we prepare augmented version of a randomly sampled subset of the CASIA-WebFace \cite{CASIA} using synthetic face images from both Masi et al.'s method \cite{masiFG17} and SREFI \cite{SREFI2}. To gauge the statistical significance of the experimental results, we sample three different sets (not necessarily subject disjoint) of 111,276 images of 1,452 subjects from CASIA-WebFace \cite{CASIA}, the ``Masi Dataset" (described in Section V of the main paper) and the Notre Dame Synthetic Face Dataset. The synthetic images from the Masi dataset are posed at [0\degree, -22\degree, -40\degree, -55\degree, -75\degree] yaw while the Notre Dame Synthetic Face Dataset consists of multiples of 30\degree increments in facial yaw for a synthetic identity. We do not explicitly check for consistency in the pose distribution in each of these datasets. As the synthetic images are graphically rendered, they are essentially masked outside the salient facial region (\emph{i.e.} zero intensity), and do not have any contextual information embedded in them.

\begin{table*}
\begin{center}
\captionsetup{justification=centering}
\caption{Effectiveness of Masi et al.'s method \cite{masiFG17} and SREFI \cite{SREFI2} as data augmentation modules, tested on IJB-B \cite{IJBB}, using sampled CASIA-WebFace (CW) \cite{CASIA} and the ResNet50 model \cite{ResNet}. We highlight the best two verification scores.}
\begin{normalsize}
\begin{tabular}{  | c | c| c| c| c|  }
\hline
{\bf Training Data} & \begin{tabular}[x]{@{}c@{}}{\bf CW \cite{CASIA} Images}\\(real identities)\end{tabular} & \begin{tabular}[x]{@{}c@{}}{\bf Synthetic Masi images}\\(real identities)\end{tabular} & \begin{tabular}[x]{@{}c@{}}{\bf Synthetic SREFI images}\\(synthetic identities)\end{tabular} & \begin{tabular}[x]{@{}c@{}}{\bf IJB-B \cite{IJBB} Performance}\\(TAR@FAR = 0.01)\end{tabular} \\
\hline
\hline
  Dataset 1  &  111,276 (1,452)  &  0  &  0  &  0.925 $\pm$ 0.04 \\
  \hline
  \hline
  Dataset 2  &  111,276 (1,452)  & 111,276 (1,452)  &  0  &  {\bf 0.939 $\pm$ 0.14} \\
  \hline
  Dataset 3  &  0  &  111,276 (1,452)  &  0  &  0.857 $\pm$ 0.18\\
  \hline
  \hline
  Dataset 4  &  111,276 (1,452)  &  0 & 111,276 (1,452)  &  {\bf 0.937 $\pm$ 0.09} \\
  \hline
  Dataset 5  &  0  &  0  &  111,276 (1,452)  &  0.835 $\pm$ 0.22\\

\hline
\end{tabular}
\label{Tab:Data_Aug_Masi_SREFI}
\end{normalsize}
\end{center}
\vspace{-0.5cm}
\end{table*}

We fine-tune the ResNet-50 network \cite{ResNet}, pre-trained on VGGFace2 \cite{VGGFace2}, separately with these datasets. The hyper-parameters, scheduling and termination policy is set the same as described in Section V of the main text. For testing each trained network we use the IJB-B verification protocol \cite{IJBB} and match templates post \emph{video} and \emph{media pooling} operations using a Pearson correlation. As shown in Table \ref{Tab:Data_Aug_Masi_SREFI}, both the augmentation methods (Datasets 2 and 4) trigger a boost in ResNet50's verification score when trained on just CASIA-WebFace (Dataset 1). Furthermore, when the number of training images and identities is kept the same, the test performance remains approximately the same. Although Masi et al. \cite{masiFG17} generated images visually look different than those generated by SREFI \cite{SREFI2}, as can be seen in Fig. 2 of the main text (forehead, jawline, prominence of 3D shape, etc.), they produce similar results when similar amount of supplemental data (\emph{i.e.} same number of images and identities) is chosen from each set. This suggests the visual appearance of the supplemental data in face recognition experiments is not as important as the size of the overall training dataset; this is consistent with recent findings in other domains \cite{Cecilia}. However, synthetic face images alone fail to capture the full range of visual diversity found in real images (Dataset 1) and consequently garner lower verification scores when used by themselves (Datasets 3 and 5) to train the ResNet50 model.

\section{Conclusion}
In this paper, we perform an extensive set of experiments to quantify the impact the underlying facial shape has on the overall appearance of a particular subject. For this work, we chose two data augmentation methods from \cite{masiFG17} and \cite{SREFI2} that use different 3D models to graphically render novel views of 2D face images (real or synthetic) with varying facial pose and shape. Utilizing these rendered images as variations of the original image (and subject) we analyze how much of a perturbation this process injects into the base identity. This analysis is performed using FID \cite{FID} as the realism metric and the ResNet50 \cite{ResNet,VGGFace2} model as an estimator of facial similarity. Since the intrinsic shape of a profile face is more visible to the network, we find the subject identity is disturbed further as we move away from a frontal pose when variable 3D models are used to render the same face image. We also observe that simply choosing 3D models based on congruence in gender and ethnicity to the base identity generates synthetic views that score similarly in realism and identity retention when compared to a strict fit-estimation function. Additionally, we quantify the effect context and background pixels have on model performance when used to train a deep network. We find the presence of such pixels in the training data, even when artificially hallucinated \cite{SREFI3}, to inject vital information to the model representation and improve its downstream authentication performance.

{\bf Reproducibility}: Most of the datasets, 3D face models and network models used in this study are publicly available at present, as listed below:
\begin{enumerate}
    \item {\bf The Notre Dame Synthetic Face Dataset \cite{SREFI2}}: \url{https://cvrl.nd.edu/projects/data/}.
    \item {\bf Gallery for ``Masi Dataset"}: shared in \cite{SREFIDonor}.
    \item {\bf Masi Renderer \cite{masiFG17}}: \url{https://github.com/iacopomasi/face_specific_augm}.
    \item {\bf ResNet50 \cite{ResNet} trained on VGGFace2 \cite{VGGFace2}}: \url{https://github.com/ox-vgg/vgg_face2}.
\end{enumerate}
The trained snapshot of the context generation GAN \cite{SREFI3}, used to hallucinate context and background in Section \ref{sec:Exp5} of the main paper, is however not publicly available and was generously shared with us by its authors for this work.

{\small
\bibliographystyle{ieee_fullname}
\bibliography{output}

\begin{thebibliography}{10}\itemsep=-1pt

\bibitem{exposeai}
Exposing.ai. available here: \url{https://exposing.ai/}.

\bibitem{RaceShape}
B. Balas and CA. Nelso.
\newblock The role of face shape and pigmentation in other-race face
  perception: an electrophysiological study.
\newblock {\em Neuropsychologia}, 48(2):498--506, 2010.

\bibitem{SREFI1}
S. Banerjee, J. Bernhard, W. Scheirer, K. Bowyer, and P. Flynn.
\newblock Srefi: Synthesis of realistic example face images.
\newblock In {\em IJCB}, 2017.

\bibitem{LEGAN}
S. Banerjee, A. Joshi, P. Mahajan, S. Bhattacharya, S. Kyal, and T. Mishra.
\newblock Legan: Disentangled manipulation of directional lighting and facial
  expressions whilst leveraging human perceptual judgements.
\newblock In {\em CVPR Workshops}, 2021.

\bibitem{SREFI2}
S. Banerjee, W. Scheirer, K. Bowyer, and P. Flynn.
\newblock Fast face image synthesis with minimal training.
\newblock In {\em WACV}, 2019.

\bibitem{SREFI3}
S. Banerjee, W. Scheirer, K. Bowyer, and P. Flynn.
\newblock On hallucinating context and background pixels from a face mask using
  multi-scale gans.
\newblock In {\em WACV}, 2020.

\bibitem{DosDonts}
A. Bansal, C. Castillo, R. Ranjan, and R. Chellappa.
\newblock The do's and don'ts for cnn-based face verification.
\newblock {\em ICCV Workshops}, 2017.

\bibitem{Marcel_BTAS}
S. Bhattacharjee, A. Mohammadi, and S. Marcel.
\newblock Spoofing deep face recognition with custom silicone masks.
\newblock In {\em BTAS}, 2018.

\bibitem{bottou2010large}
L. Bottou.
\newblock Large-scale machine learning with stochastic gradient descent.
\newblock In {\em COMPSTAT}. 2010.

\bibitem{ExternalFeatImp}
C. Brown, E. Portch, FC. Skelton, C. Fodarella, H. Kuivaniemi-Smith, K. Herold,
  PJB. Hancock, and CD. Frowd.
\newblock The impact of external facial features on the construction of facial
  composites.
\newblock {\em Ergonomics}, 62(4):575--592, 2019.

\bibitem{PoggioShape}
R. Brunelli and T. Poggio.
\newblock Face recognition: features versus templates.
\newblock {\em IEEE Trans. on Pattern Analysis and Machine Intelligence},
  15(10):1042--1052, 1993.

\bibitem{celeb500k}
J. Cao, Y. Li, and Z. Zhang.
\newblock Celeb-500k: A large training dataset for face recognition.
\newblock In {\em ICIP}, 2018.

\bibitem{VGGFace2}
Q. Cao, L. Shen, W. Xie, O.~M. Parkhi, and A. Zisserman.
\newblock Vggface2: A dataset for recognizing faces across pose and age.
\newblock In {\em arXiv:1710.08092}.

\bibitem{starganv2}
Y. Choi, Y. Uh, J. Yoo, and JW. Ha.
\newblock Stargan v2: Diverse image synthesis for multiple domains.
\newblock In {\em CVPR}, 2020.

\bibitem{ASM}
TF. Cootes, CJ. Taylor, DH. Cooper, and J. Graham.
\newblock Active shape models – their training and application.
\newblock {\em CVIU}, 61:38--59, 1995.

\bibitem{arcface}
J. Deng, J. Guo, N. Xue, and S. Zafeiriou.
\newblock Arcface: Additive angular margin loss for deep face recognition.
\newblock In {\em CVPR}. 2019.

\bibitem{lwc_iccvw}
J. Deng, J. Guo, D. Zhang, Y. Deng, X. Lu, and S. Shi.
\newblock Lightweight face recognition challenge.
\newblock In {\em ICCV Workshops}, 2019.

\bibitem{discofacegan}
Y. Deng, J. Yang, D. Chen, F. Wen, and X. Tong.
\newblock Disentangled and controllable face image generation via 3d
  imitative-contrastive learning.
\newblock In {\em CVPR}, 2020.

\bibitem{InfluenceFacialFeature}
JA. Diego-Mas, F. Fuentes-Hurtado, V. Naranjo, and M. Alcañiz.
\newblock The influence of each facial feature on how we perceive and interpret
  human faces.
\newblock {\em i-Perception}, 2020.

\bibitem{StatShape}
IL. Dryden and KV. Mardia.
\newblock Statistical shape analysis.
\newblock {\em John Wiley}, 1998.

\bibitem{Context1}
C. Frowd, V. Bruce, A. McIntyre, and P. Hancock.
\newblock The relative importance of external and internal features of facial
  composites.
\newblock {\em British Journal of Psychology}, 98(1):61--77, 2007.

\bibitem{Context2}
C. Frowd, W. Erickson, J. Lampinen, F. Skelton, A. McIntyre, and P. Hancock.
\newblock A decade of evolving composite techniques: Regression-and
  meta-analysis.
\newblock {\em Journal of Forensic Practice}, 2015.

\bibitem{HairImp}
C. Frowd and G. Hepton.
\newblock The benefit of hair for the construction of facial composite images.
\newblock {\em The British Journal of Forensic Practice}, 2009.

\bibitem{MSCeleb}
Y. Guo, L. Zhang, Y. Hu, X. He, and J. Gao.
\newblock Ms-celeb-1m: A dataset and benchmark for large-scale face
  recognition.
\newblock In {\em ECCV}, 2016.

\bibitem{HassFront}
T. Hassner, S. Harel, E. Paz, and R. Enbar.
\newblock Effective face frontalization in unconstrained images.
\newblock In {\em CVPR}, 2015.

\bibitem{ResNet}
K. He, X. Zhang, S. Ren, and J. Sun.
\newblock Deep residual learning for image recognition.
\newblock {\em CVPR}, 2016.

\bibitem{FID}
M. Heusel, H. Ramsauer, T. Unterthiner, B. Nessler, and S. Hochreiter.
\newblock Gans trained by a two time-scale update rule converge to a local nash
  equilibrium.
\newblock In {\em NeurIPS}, 2017.

\bibitem{lfw}
G.~B. Huang, M. Ramesh, T. Berg, and E. Learned-Miller.
\newblock Labeled faces in the wild: A database for studying face recognition
  in unconstrained environments.
\newblock In {\em Technical Report 07--49}, 2007.

\bibitem{stylegan2}
T. Karras, S. Laine, M. Aittala, J. Hellsten, J. Lehtinen, and T. Aila.
\newblock Analyzing and improving the image quality of {StyleGAN}.
\newblock In {\em CVPR}, 2020.

\bibitem{MegaFace}
I. Kemelmacher-Shlizerman, S. Seitz, D. Miller, and E. Brossard.
\newblock The megaface benchmark: 1 million faces for recognition at scale.
\newblock In {\em CVPR}, 2016.

\bibitem{Dlib}
D.~E. King.
\newblock Dlib-ml: A machine learning toolkit.
\newblock In {\em JMLR}, volume~10, pages 1755--1758, 2009.

\bibitem{DLNature}
Y. LeCun, Y. Bengio, and G. Hinton.
\newblock Deep learning.
\newblock {\em Nature}, 521(7553):436--444, 2015.

\bibitem{CAERNet}
J. Lee, S. Kim, S. Kim, J. Park, and K. Sohn.
\newblock Context-aware emotion recognition networks.
\newblock {\em ICCV}, 2019.

\bibitem{masiFG17}
I. Masi, T. Hassner, A.~T. Tran, and G. Medioni.
\newblock Rapid synthesis of massive face sets for improved face recognition.
\newblock {\em FG}, 2017.

\bibitem{masi_ijcv}
I. Masi, AT. Tran, T. Hassner, G. Sahin, and G. Medioni.
\newblock Face-specific data augmentation for unconstrained face recognition.
\newblock {\em IJCV}, 127:642--667, 2019.

\bibitem{MasiAug}
I. Masi, A.~T. Tran, J.~T. Leksut, T. Hassner, and G. Medioni.
\newblock Do we really need to collect millions of faces for effective face
  recognition?
\newblock In {\em ECCV}, 2016.

\bibitem{ijbc}
B. Maze and et al.
\newblock Iarpa janus benchmark–c: Face dataset and protocol.
\newblock In {\em ICB}, 2018.

\bibitem{FGNet}
G. Panis, A. Lanitis, N. Tsapatsoulis, and T. Cootes.
\newblock Overview of research on facial ageing using the fg-net ageing
  database.
\newblock {\em IET Biometrics}, 5(2):37--46, 2016.

\bibitem{VGG}
O.~M. Parkhi, A. Vedaldi, and A. Zisserman.
\newblock Deep face recognition.
\newblock In {\em BMVC}, 2015.

\bibitem{Basel}
P. Paysan, R. Knothe, B. Amberg, S. Romdhani, and T. Vetter.
\newblock A 3d face model for pose and illumination invariant face recognition.
\newblock In {\em AVSS}, 2009.

\bibitem{Lessons1000}
K. Peng, A. Mathur, and A. Narayanan.
\newblock Mitigating dataset harms requires stewardship: Lessons from 1000
  papers.
\newblock {\em arXiv:2108.02922}.

\bibitem{PJP_FG}
P.J. Phillips.
\newblock A cross benchmark assessment of a deep convolutional neural network
  for face recognition.
\newblock In {\em FG}, 2017.

\bibitem{phillips2005overview}
P.J. Phillips, P. Flynn, T. Scruggs, K. Bowyer, J. Chang, K. Hoffman, J.
  Marques, J. Min, and W. Worek.
\newblock Overview of the face recognition grand challenge.
\newblock In {\em CVPR}, 2005.

\bibitem{SREFIDonor}
P.~J. Phillips, P. Flynn, and K. Bowyer.
\newblock Lessons from collecting a million biometric samples.
\newblock {\em IVC}, 2016.

\bibitem{Marcel_IWBF}
R. Ramachandra, S. Venkatesh, KB. Raja, S. Bhattacharjee, P. Wasnik, S. Marcel,
  and C. Busch.
\newblock Custom silicone face masks: Vulnerability of commercial face
  recognition systems \& presentation attack detection.
\newblock In {\em IWBF}, 2019.

\bibitem{Context3}
A. Rice, P.J. Phillips, V. Natu, X. An, and A.J. O'Toole.
\newblock Unaware person recognition from the body when face identification
  fails.
\newblock {\em Psychological Science}, 24:2235--2243, 2013.

\bibitem{Google_FaceNet}
F. Schroff, D. Kalenichenko, and J. Philbin.
\newblock Facenet: A unified embedding for face recognition and clustering.
\newblock In {\em CVPR}, 2015.

\bibitem{SurfaceCue}
M. Sormaz, AW. Young, and TJ. Andrews.
\newblock Contributions of feature shapes and surface cues to the recognition
  of facial expressions.
\newblock {\em Vision Research}, 127:1--10, 2016.

\bibitem{Cecilia}
C. Summers and M.J. Dinneen.
\newblock Improved mixed-example data augmentation.
\newblock In {\em WACV}, 2019.

\bibitem{Facebook_Deepface}
Y. Taigman, M. Yang, M. Ranzato, and L. Wolf.
\newblock Deepface: Closing the gap to human-level performance in face
  verification.
\newblock In {\em CVPR}, 2014.

\bibitem{ComprehensiveStudy}
P. Terhörst, JN. Kolf, M. Huber, F. Kirchbuchner, N. Damer, A. Morales, J.
  Fierrez, and A. Kuijper.
\newblock A comprehensive study on face recognition biases beyond demographics.
\newblock In {\em arXiv:2103.01592}, 2021.

\bibitem{Tinsley_2021_WACV}
P. Tinsley, A. Czajka, and P. Flynn.
\newblock This face does not exist... but it might be yours! identity leakage
  in generative models.
\newblock In {\em WACV}, 2021.

\bibitem{deepFR}
M. Wang and W. Deng.
\newblock Deep face recognition: A survey.
\newblock {\em Neurocomputing}, 429:215--244, 2021.

\bibitem{FRAugSurvey}
X. Wang, K. Wang, and S. Lian.
\newblock A survey on face data augmentation for the training of deep neural
  networks.
\newblock {\em Neural Computing and Applications}, 32:15503--15531, 2020.

\bibitem{Shape_BodyWeight}
L. Wen, G. Guo, and X. Li.
\newblock A study on the influence of body weight changes on face recognition.
\newblock In {\em IJCB}, 2014.

\bibitem{IJBB}
C. Whitelam, E. Taborsky, A. Blanton, B. Maze, J. Adams, T. Miller, N. Kalka,
  A.~K. Jain, J.~A. Duncan, K. Allen, J. Cheney, and P. Grother.
\newblock Iarpa janus benchmark-b face dataset.
\newblock In {\em CVPR Workshops}, 2017.

\bibitem{LightCNN}
X. Wu, R. He, Z. Sun, and T. Tan.
\newblock A light cnn for deep face representation with noisy labels.
\newblock {\em arXiv:1511.02683}, 2017.

\bibitem{CASIA}
D. Yi, Z. Lei, S. Liao, and S.~Z. Li.
\newblock Learning face representation from scratch.
\newblock In {\em arXiv:1411.7923}.

\end{thebibliography}
}

\end{document}